\pdfoutput=1

\documentclass[11pt]{article}

\usepackage[]{acl}

\usepackage{times}
\usepackage{latexsym}
\usepackage{graphicx}
\usepackage{booktabs,tabularx,enumitem,ragged2e}
\usepackage{times}
\usepackage{latexsym}
\usepackage{multirow}
\usepackage{xcolor}
\usepackage{subfigure}
\usepackage{amsmath,amsthm,amssymb}
\usepackage[boldmath]{numprint}
\newcommand{\new}[1]{\textcolor{black}{{#1}}}

\newcommand{\simi}{\operatorname{sim}}

\usepackage[T1]{fontenc}

\usepackage[utf8]{inputenc}

\usepackage{microtype}

\title{Explanation Graph Generation via Pre-trained Language Models: \\ An Empirical Study with Contrastive Learning}

\author{Swarnadeep Saha \quad Prateek Yadav \quad Mohit Bansal
\\ 
  UNC Chapel Hill\\ 
  \texttt{\{swarna, prateek, mbansal\}@cs.unc.edu}
}

\begin{document}
\maketitle
\begin{abstract}
Pre-trained sequence-to-sequence language models have led to widespread success in many natural language generation tasks. However, there has been relatively less work on analyzing their ability to generate structured outputs such as graphs. Unlike natural language, graphs have distinct structural and semantic properties in the context of a downstream NLP task, e.g., generating a graph that is connected and acyclic can be attributed to its structural constraints, while the semantics of a graph can refer to how meaningfully an edge represents the relation between two node concepts. In this work, we study pre-trained language models that generate explanation graphs in an end-to-end manner and analyze their ability to learn the structural constraints and semantics of such graphs. We first show that with limited supervision, pre-trained language models often generate graphs that either violate these constraints or are semantically incoherent. Since curating large amount of human-annotated graphs is expensive and tedious, we propose simple yet effective ways of graph perturbations via node and edge edit operations that lead to structurally and semantically positive and negative graphs. Next, we leverage these graphs in different contrastive learning models with Max-Margin and InfoNCE losses. Our methods lead to significant improvements in both structural and semantic accuracy of explanation graphs and also generalize to other similar graph generation tasks. Lastly, we show that human errors are the best negatives for contrastive learning and also that automatically generating more such human-like negative graphs can lead to further improvements.\footnote{\new{Our code and models are publicly available at \url{https://github.com/swarnaHub/ExplagraphGen}.}}

\end{abstract}

\begin{figure}[tbh!]
\centering
    \includegraphics[width=0.92\columnwidth]{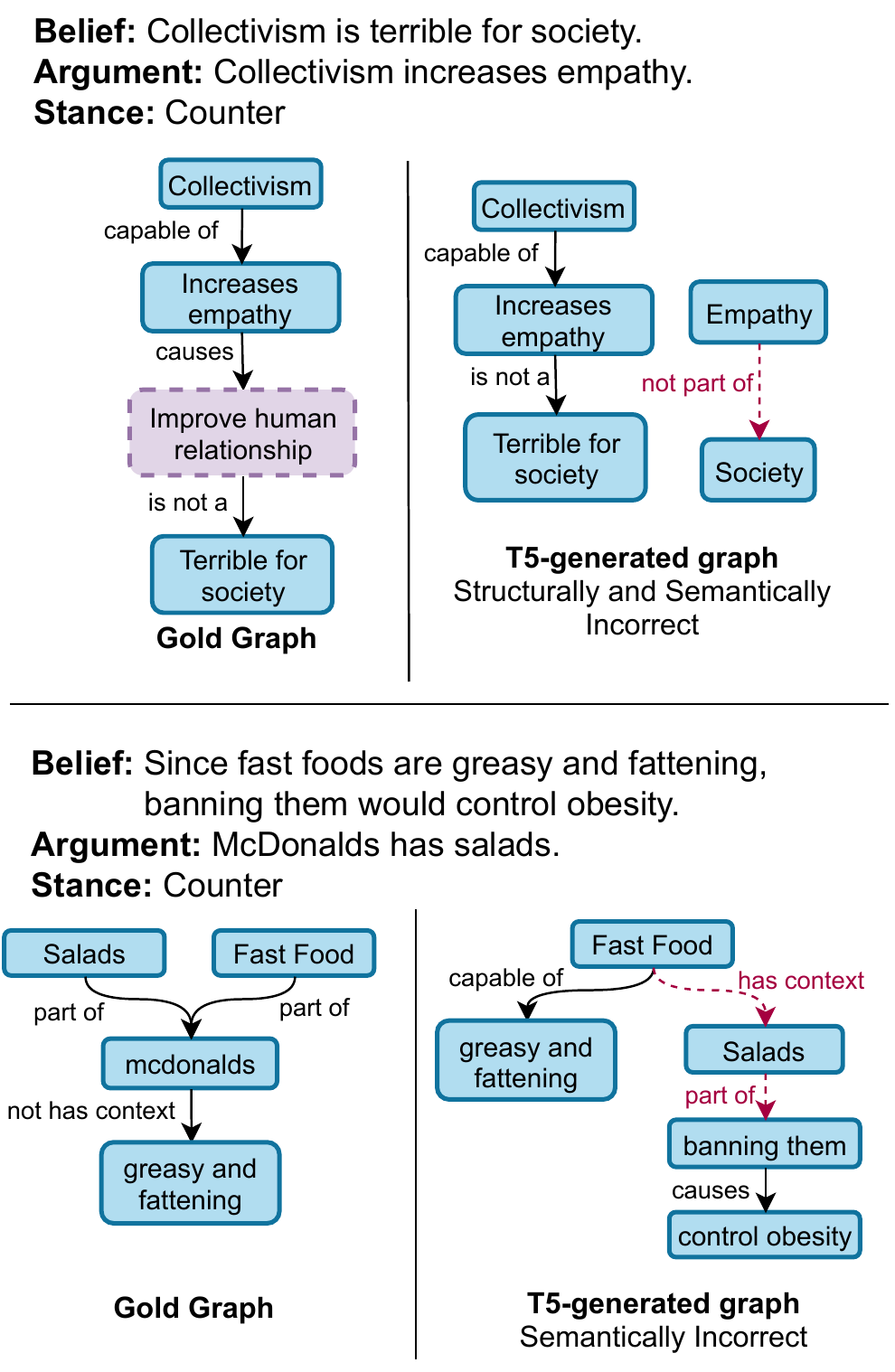}
    \vspace{-7pt}
    \caption{Two representative examples from ExplaGraphs \cite{saha2021explagraphs} showing the belief, argument, stance, gold explanation graph, and T5-generated explanation graph. The dashed nodes represent commonsense nodes and the dashed edges are incorrect edges. The first generated graph is structurally incorrect and the second graph is semantically incorrect.\label{fig:intro_fig}
    }
    \vspace{-10pt}
\end{figure}

\section{Introduction}

Pre-trained sequence-to-sequence language models (PLMs) like BART~\cite{lewis2020bart} and T5~\cite{2020t5} have led to significant advances in many natural language generation tasks like text summarization and machine translation. The models are pre-trained on massive amounts of text data with self-supervision, thus enabling them to construct coherent natural language sentences for downstream tasks. This then raises the question whether pre-trained language models, trained on free-form natural language data, can also adapt themselves to generate structured outputs like graphs. Graphs are common in NLP tasks that involve representing structured knowledge in the form of knowledge bases~\cite{guarino1995ontologies}, constructing event chains from documents~\cite{chambers2009unsupervised}, or more recent work on encoding reasoning chains, explanations, or deductive proofs~\cite{saha2020prover, tafjord2020proofwriter, dalvi2021explaining}.

Graphs differ from free-form natural language. In the context of NLP, natural language graphs (consisting of textual nodes and edges) can have distinct structural and semantic properties. For example, consider a recently proposed commonsense explanation graph generation task shown in Fig.~\ref{fig:intro_fig}~\cite{saha2021explagraphs}. Each example shows a belief, an argument and an explanation graph explaining how the argument supports or refutes the belief. \new{These explanation graphs encode structured knowledge (augmented with commonsense) and consist of concepts as nodes and relations from ConceptNet \cite{liu2004conceptnet} as edges. For example, the second graph encodes the knowledge that ``both salads and fast food are part of mcdonalds and hence mcdonalds is not greasy and fattening'', thus explicitly refuting the belief.} From prior work, the structural constraints enforce the graphs to be connected directed acyclic and the nodes to contain at least two concepts from the belief and two from the argument. The semantic aspect deals with commonsense and evaluates whether each edge expresses coherent relational knowledge and if the whole graph explains the stance.

Following~\citet{saha2021explagraphs}, we represent graphs as strings composed of concatenated edges and fine-tune T5 to generate graphs in an autoregressive manner. We observe that while moderate amount of supervision enables the model to learn valid graph encodings, the graphs frequently violate task-specific structural constraints (like connectivity). For instance, the first example in Fig.~\ref{fig:intro_fig} shows a graph generated by T5 that is disconnected and hence structurally incorrect. Moreover, for the fraction of graphs that are structurally correct, the model also makes commonsense mistakes, a type of semantic error, by inferring wrong or incoherent relations between concepts. Both T5-generated graphs shown in Fig.~\ref{fig:intro_fig} contain incoherent or non-commonsensical edges (marked by dashed arrows) like ``fast food; has context; salads''. Based on these observations, we study PLMs that generate explanation graphs in an end-to-end manner \new{and analyze their ability to learn the structural constraints as well as the semantics of such graphs.}

While a general recipe towards improving the structural and semantic aspects of graph generation can be via large-scale training with more human-annotated graphs, it is prohibitive under most practical scenarios because of the cognitive load associated with a complex data creation task like graph annotation~\cite{dalvi2021explaining, saha2021explagraphs}. Hence, we propose simple yet effective methods of graph perturbations that perform various kinds of node and edge addition, deletion, and replacement operations to construct \new{structurally} and \new{semantically} positive (correct) and negative (incorrect) graphs. Overall, we leverage three types of negative graphs (synthetic \new{structural}, synthetic semantic, and human-created semantic) and develop multiple contrastive learning models~\cite{hjelm2018learning, chen2020simple, khosla2020supervised, gunel2020supervised} for effectively distinguishing between correct and incorrect graphs. Our first method is a \textit{Generate-and-Refine} model that first generates an initial graph and further refines it using another T5 model. Next, we propose two improved models -- one that uses the negative graphs in a max-margin formulation and another that uses both positive and negative graphs with a InfoNCE~\cite{Oord2018RepresentationLW} contrastive loss.  On two real-world tasks of explanation graph generation and temporal graph generation, with varied node and edge semantics, we observe that our proposed methods and graph perturbation techniques generalize well and lead to improvements in both structural and semantic accuracy of graphs. \new{Further analysis of different types of negative graphs reveal that the human-error graphs are the hardest, most diverse, and hence the best type of negatives to learn from in contrastive learning. Hence, we also develop methods to automatically generate more such human-like semantic negative graphs, which leads to further improvements.} We summarize our contributions as follows.

\begin{itemize}[nosep, wide=0pt, leftmargin=*, after=\strut]
    \item \new{We present a detailed analysis of graph structure and semantics for end-to-end explanation graph generation via pre-trained language models.}
    \item We propose simple yet effective graph perturbation techniques for constructing positive and negative graphs (incl. human errors) and use them in different graph contrastive learning models.
    \item Our methods lead to significant improvements in both structural and semantic accuracy of explanation graphs and also generalize to other similar (e.g., temporal) graph generation tasks.
\end{itemize}{}

\section{Related Work}
\noindent \textbf{Graph Generation from Language Models.} Representative works on graph generation from language models include knowledge graph completion models like Comet~\cite{bosselut2019comet, hwang2021comet} that fine-tune GPT~\cite{radford2019language, brown2020language} and BART~\cite{lewis2020bart}, generation of event influence graphs~\cite{tandon2019wiqa, madaan2020eigen}, partially ordered scripts~\cite{sakaguchi2021proscript}, temporal graphs~\cite{madaan2021neural}, entailment trees~\cite{dalvi2021explaining}, proof graphs~\cite{saha2020prover, tafjord2020proofwriter, saha2021multiprover} and commonsense explanation graphs~\cite{saha2021explagraphs}. \new{Linguistic tasks like syntactic parsing \cite{zhou-etal-2020-limit, mohammadshahi2021recursive, kondratyuk201975} and semantic parsing \cite{chen2020low, shin2021constrained} have also made use of language models.} There is also a large body of work on building generative models for learning unconditional graph distributions~\cite{you2018graphrnn, simonovsky2018graphvae, grover2019graphite, liao2019efficient, shi2020graphaf} without any semantics attached to the graphs. Our novelty lies in presenting the first systematic analysis of \new{structure} and semantics of graph generation for two downstream NLP tasks using pre-trained language models and improving them via constrastive learning.

\noindent \textbf{Data Augmentation and Contrastive Learning.} \new{Data Augmentation for NLP \cite{hedderich2020survey, feng2021survey, chen2021empirical} has been a powerful tool in low-data settings, ranging from its early usages with synonym replacement \cite{kolomiyets2011model, wang2015s} to more recent methods of perturbing hidden representations \cite{miyato2016adversarial, shen2020simple}}. Contrastive learning, beyond its historical use in learning robust image representations~\cite{chopra2005learning, hadsell2006dimensionality, gutmann2010noise, hoffer2015deep, hjelm2018learning, chen2020simple, he2020momentum} has been explored in supervised scenarios~\cite{khosla2020supervised, gunel2020supervised} and for NLP, in training self-supervised language models~\cite{fang2020cert}, learning sentence representations~\cite{gao2021simcse}, document clustering~\cite{zhang2021supporting}, summarization~\cite{liu2021simcls, cao2021cliff} and generic text generation~\cite{lee2020contrastive}. It has also been used in unconditional graph representation learning~\cite{you2020graph, hassani2020contrastive, zhu2021graph}. We follow this rich line of work to explore their applicability in supervised graph generation tasks from pre-trained language models in low-resource settings.

\noindent \textbf{Generative Commonsense Reasoning.} While traditional commonsense reasoning tasks are discriminative in nature~\cite{zellers2018swag, talmor2019commonsenseqa, sap2019socialiqa, bisk2020piqa, sakaguchi2020winogrande, talmor2021commonsenseqa}, recent focus on generative evaluation have led to the development of tasks and benchmarks that explore unstructured commonsense sentence generation~\cite{lin2020commongen}, event influence graph generation~\cite{madaan2020eigen}, commonsense explanation graph generation~\cite{saha2021explagraphs}, etc. We experiment with two graph generation tasks, primarily focusing on ExplaGraphs \cite{saha2021explagraphs} because of the clear distinction in the underlying structural constraints and the semantic aspect dealing with commonsense.

\section{Motivation and Background}

\new{Our primary task of interest is a recently proposed commonsense explanation graph generation task called ExplaGraphs~\cite{saha2021explagraphs}. In Sec.~\ref{sec:temporal}, we also experiment with another related task of temporal graph generation \cite{madaan2020eigen}. In both these tasks, the structural aspect deals with satisfying certain task-specific constraints on the graph (like connectivity) and the semantic aspect deals with the construction of meaningful edges (that adhere to commonsense). Below we discuss ExplaGraphs briefly and analyze pre-trained language models for their ability to generate explanation graphs.}

\paragraph{ExplaGraphs \cite{saha2021explagraphs}. }In this task, given a belief and an argument, an agent has to perform two sub-tasks -- predict the stance (support/counter) and also generate an explanation graph explaining the stance. \new{Explanation graphs are structured explanations that capture explicit reasoning chains between the belief and the argument, thereby making models more interpretable.} Formally, an explanation graph is a connected DAG with nodes as concepts and edges as commonsense relations between two concepts (See Fig.~\ref{fig:intro_fig}). \new{The concepts are either part of the belief or the argument (represented with solid boxes) or any external commonsense phrase (represented with dashed boxes). Each edge in the graph forms a coherent sentence and the graph, when read as a whole, forms reasoning structures explaining why the argument supports or refutes the belief.} \citet{saha2021explagraphs} evaluate explanation graphs by defining two accuracy metrics -- (1) \emph{Structural Correctness Accuracy (StCA)}: Fraction of graphs that satisfy all structural constraints, and (2) \emph{Semantic Correctness Accuracy (SeCA)}: Fraction of graphs that are both structurally and semantically correct. A graph is considered structurally correct if it satisfies the following constraints: (1) it is connected, (2) it is a DAG, (3) the edge relations belong to a pre-defined list, (4) there are at least two concepts from the belief and two from the argument.  If all these constraints are satisfied, the graph is next evaluated for semantic correctness by a model-based metric~\cite{saha2021explagraphs}. It works on the principle that an explanation graph is semantically correct if the stance inferred from the belief and the graph matches the gold stance. Refer to Appendix~\ref{appendix:metrics} for a detailed description of all evaluation metrics.

\paragraph{Baseline T5 Model. }Following prior work~\cite{saha2021explagraphs}, we generate explanation graphs as post-hoc explanations by conditioning on the belief, argument and the predicted stance.\footnote{These are rationalizing models~\cite{rajani2019explain, hase2020leakage} that first predict the stance, followed by the graph. While graphs can also be generated first, followed by the stance, we experiment with one model family for this work.} The stance prediction model is a fine-tuned RoBERTa model~\cite{liu2019roberta} which we keep unaltered from prior work and focus on the graph generation sub-task. We generate graphs as linearized strings in an end-to-end manner by leveraging an encoder-decoder pre-trained language model, T5~\cite{2020t5}. The input to the model is the concatenated belief, argument and the stance along with a prefix \textit{``Generate an Explanation Graph for''}. The graphs are encoded as concatenated bracketed edges, in which the edges are ordered according to the Depth First Search (DFS) order of the nodes. While we choose T5 because of its superior performance~\cite{saha2021explagraphs}, we do not make any model-specific assumptions and graphs can be generated via any encoder-decoder style pre-trained language model (\new{e.g., see Appendix~\ref{appendix:results} for results with BART)}.

\paragraph{Analysis of T5 Baseline.} We analyze the quality of the explanation graphs generated by T5 in Table~\ref{tab:data_size}. We vary the amount of training data from 500 to 2368 samples (all) and report StCA and SeCA along with other metrics like Graph-BertScore (G-BS) introduced in prior work~\cite{saha2021explagraphs}. While the structural accuracy improves with increase in training data, the gain saturates quickly and even after training on the entire data, we find a significant fraction of graphs to violate the structural constraints. We note that a high 91\% of T5's generations are valid graph encodings i.e., the generated strings can be parsed into graphical structures (without any post-processing), suggesting that T5 is able to learn the graph encoding from a fairly small amount of supervision. However, it fails to satisfy the various structural constraints -- (1) 20\% of the graphs are disconnected, (2) 6\% of the graphs contain cycles, and (3) 14\% of the graphs have less than two concepts from the belief or from the argument. Note that these constraints are not encoded in the model, thus making them fairly hard to learn from limited supervision. On the fraction of structurally correct graphs, the model makes further semantic errors and a lower SeCA of 35\% demonstrates that. In Fig.~\ref{fig:intro_fig}, we show examples of structurally incorrect and semantically incorrect graphs generated by T5. Overall, these results indicate that there is a significant scope for improvement both on graph structure and semantics, thus motivating us to develop methods with design choices aimed at improving both aspects.

\begin{table}[t]
\small
\centering
\begin{tabular}{rrrrrr}
\toprule
Count & StCA$\uparrow$ & SeCA$\uparrow$ & G-BS$\uparrow$ & GED$\downarrow$ & EA$\uparrow$ \\ \midrule
500 & 42.5 & 20.7 & 36.3 & 0.68 & 20.4 \\
1000 & 49.2 & 23.7 & 42.2 & 0.63 & 26.2 \\
1500 & 50.7 & 33.2 & 43.4 & 0.61 & 28.2 \\
2368 & 51.0 & 34.7 & 43.9 & 0.61 & 29.5 \\
\bottomrule
\end{tabular}
\vspace{-5pt}
\caption{Performance of T5-large with varying amount of training data on ExplaGraphs test set.
}
\vspace{-15pt}
\label{tab:data_size}
\end{table}

\section{Graph Perturbations}
\label{sec:data_augmentation}

Most prior works that collect human-annotated graphs for a \new{downstream} NLP task have found such collection processes to be quite expensive and tedious~\cite{tandon2019wiqa, dalvi2021explaining, saha2021explagraphs}. For instance,~\citet{saha2021explagraphs} obtained high-quality data only after multiple rounds of refinement and~\citet{dalvi2021explaining} employ trained expert annotators for entailment tree construction. The corresponding datasets are also relatively small in size (2-3k), thus limiting the prospect of large-scale training. Hence, our approach towards improving explanation graph generation is through data augmentation techniques that perturb human-curated graphs to construct positive and negative graphs. As noted earlier, we wish to construct graphs that enable better learning of structural graph constraints and their semantics.

\subsection{Positive Graph Perturbations} 
\begin{figure}[t]
    \centering
    \includegraphics[width=0.95\columnwidth]{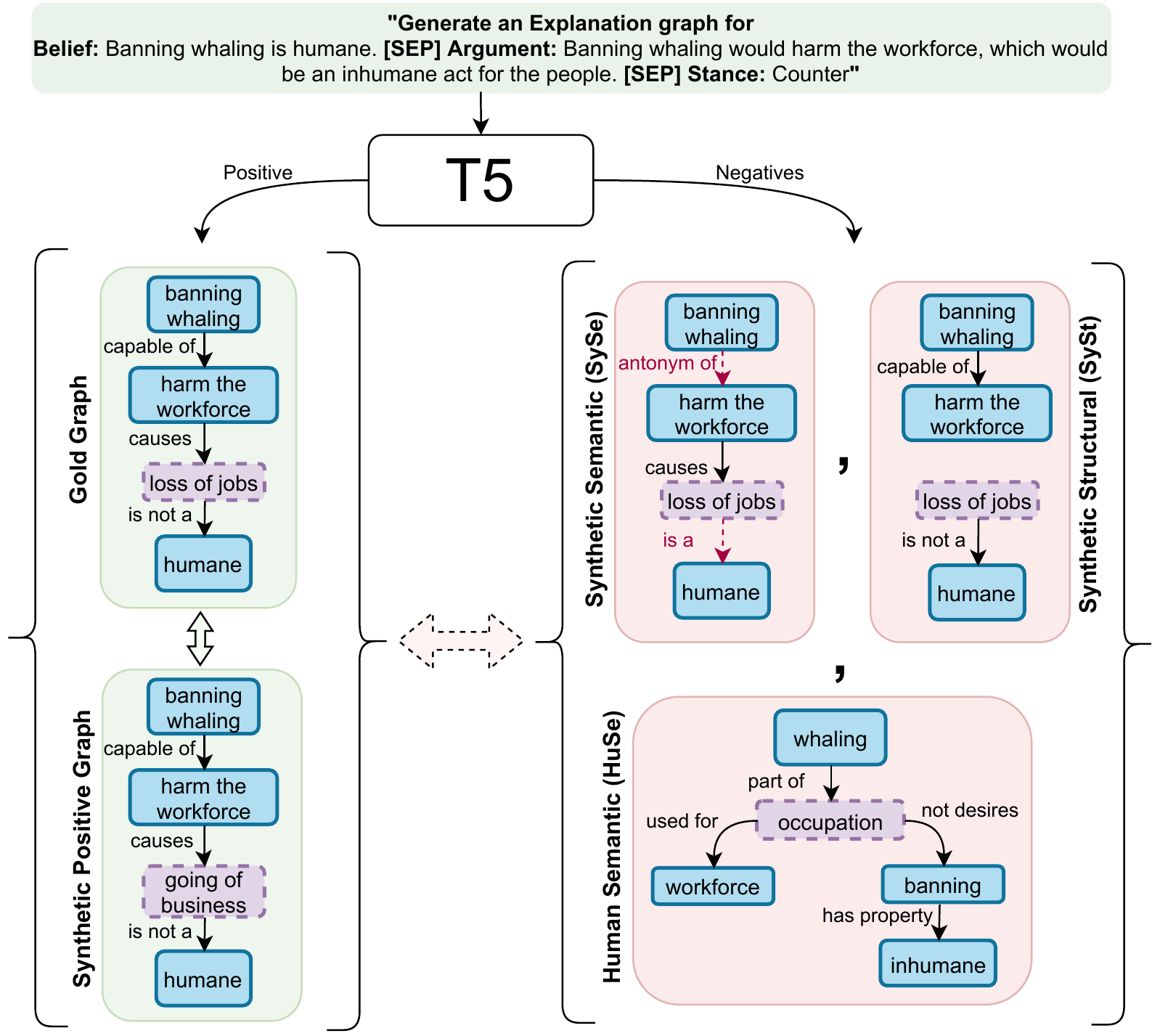}
    \vspace{-5pt}
    \caption{Our T5-based contrastive learning framework for graph generation using positively and three kinds of negatively perturbed graphs.}
    \label{fig:data_aug}
    \vspace{-15pt}
\end{figure}

One simple method to augment existing training data is to create synthetic positive graphs. These graphs should be created such that all the task-specific constraints continue to hold upon perturbations. E.g., removing a node that makes the graph disconnected is a prohibitive action. Hence, we choose nodes (concepts) that are not part of the belief or the argument (also termed as commonsense nodes) and replace them with phrases that are synonymous to the original phrases. To do so, we select words from the concept with POS tags of Adjective, Noun, Adverb, or Verb and replace them with that synonym from Wordnet \cite{miller1995wordnet} for which the cosine similarity of their word2vec representations \cite{mikolov2013efficient} is the highest.\footnote{We also tried similar replacement operations with antonyms. However, they often lead to semantically inconsistent graphs. E.g., \textit{A causes B} does not always imply \textit{A not causes not B} or \textit{not A not causes not B.}} Fig. \ref{fig:data_aug} shows an example of a positive graph perturbation where the node ``loss of jobs'' is replaced with ``going of business''. Note that our node replacement operations will always lead to structurally similar graphs. Automatically constructing structurally diverse positive graphs is a challenging problem and we leave that for future work.

\subsection{Negative Graph Perturbations}
In order to enable the model to learn from explicit hard negatives, we construct three diverse types of graphs -- synthetically constructed structural negatives for learning graph constraints and synthetic and human-created semantic negatives to capture a fairly large space of semantically incorrect graphs. Below we discuss the construction of these graphs.

\paragraph{Synthetic \& Structurally Negative Graphs (SySt).} As shown previously, one common source of errors in the generated explanation graphs is the violation of structural constraints. To enable learning these constraints, we generate four types of negative graphs by performing the following perturbations on each ground-truth graph: (1) removing an edge at random such that the resultant graph becomes disconnected, (2) adding an edge between two randomly chosen nodes such that the resultant graph becomes cyclic, (3) adding and removing one edge at random such that the resultant graph becomes both disconnected and cyclic, (4) removing a node randomly such that the resultant graph contains less than two concepts from the belief or argument. Fig.~\ref{fig:data_aug} shows an example of a disconnected graph created as part of the structurally negative graphs.

\paragraph{Synthetic \& Semantic Negative Graphs (SySe).} We also construct semantically incorrect negative explanation graphs. While the previous category of negative graphs (SySt) captures structural constraints, SySe captures the relational knowledge in graphs. Semantic incorrectness typically arises from inappropriate relations that do not adhere to human commonsense (``loss of jobs; is a; humane''). We create such negative graphs by selecting a random number of edges and then replacing the relations with some other relations. Fig.~\ref{fig:data_aug} shows a semantic negative graph in which the relations marked \new{with dashed lines} are perturbed.

\paragraph{Human-created \& Semantic Negative Graphs (HuSe). }The space of semantically incorrect graphs is fairly large and in order to augment our synthetic negative graphs with harder structurally-diverse negatives, we make use of human-created incorrect graphs \new{from prior work \cite{saha2021explagraphs}}.\footnote{\new{Publicly released by~\citet{saha2021explagraphs} at \url{https://github.com/swarnaHub/ExplaGraphs/blob/main/data/refinement_graphs_train.tsv}.}} \new{Humans make subtle errors, thus making them ideal negative candidates for contrastive learning.} ExplaGraphs was constructed via an iterative framework in which the graphs are iteratively refined (up to two times) until they are verified as correct. We treat these refined graphs as negatives. Specifically, in two rounds, if an initial graph $\mathcal{G}_1$ is refined into graphs $\mathcal{G}_2$ and $\mathcal{G}_3$ successively, then $\mathcal{G}_1$ and $\mathcal{G}_2$ are considered as negative graphs. Unlike SySe which only perturb the relations, these negatives are structurally diverse (see Fig.~\ref{fig:data_aug}) and capture semantics not just at the level of each edge but for the graph as a whole (e.g., a graph might be refined because it does not explain the stance). Note that human-created graphs can only be semantically incorrect, since their structural correctness is already ensured during construction.

\section{Augmentation with Perturbed Graphs}
\label{sec:model}
Next we propose different methods of leveraging these positive and negative graphs for explanation graph generation. Our models either use only positive graphs as simple data augmentation, only negative graphs in a max-margin model, or both in a \textit{Generate \& Refine} model and a Contrastive model.

\subsection{Augmentation with Positive Graphs}
\label{sec:positive}
In this first simple approach, we augment the training data with the synthetically created positive graphs and retrain the baseline T5 model. 

\subsection{Max-Margin Graph Generation Model}
\label{sec:max-margin}
Our next model leverages the negatively perturbed graphs in a max-margin formulation. During training, given a (belief, argument, stance) context $x$, a ground truth graph $\mathcal{G}^{(g)}$ and a negative graph $\mathcal{G}^{(n)}$, linearized into a sequence of words $\{y_i^{(g)}\}_{i=1}^k$ and $\{y_i^{(n)}\}_{i=1}^l$ respectively, we define the loss function $\mathcal{L}$ as a linear combination of the standard cross-entropy loss $\mathcal{L}_{\mathit{CE}}$ and a max-margin loss $\mathcal{L}_{\mathit{MM}}$, defined between a word $y_i^{(g)}$ of the positive graph and a word $y_i^{(n)}$ of the negative graph.
\begin{eqnarray}
    \mathcal{L}_{\mathit{CE}} = \sum_{i} - log P_\theta(y_i^{(g)}|y_{<i}^{(g)}, x) \nonumber\\
    \mathcal{L}_{\mathit{MM}} = \sum_{i} \max(0, log P_\theta(y_i^{(g)}|y_{<i}^{(g)}, x) \nonumber \\
    - \log P_\theta(y_i^{(n)}|y_{<i}^{(n)}, x)  + \beta) \nonumber \\
    \mathcal{L} = \mathcal{L}_{\mathit{CE}} + \alpha \mathcal{L}_{\mathit{MM}} \nonumber
\end{eqnarray}
where $\alpha$ and $\beta$ (margin) are hyperparameters. As noted earlier, the baseline model often makes commonsense mistakes in distinguishing between positive and negative relations (``causes'' vs ``not causes'') and our relation perturbing negative graphs and the max-margin loss component facilitate learning a better boundary between them.

\subsection{Generate \& Refine Graph Generation}
\label{sec:gen-refine}
ExplaGraphs was constructed using a ``Refinement'' phase wherein the initially constructed graphs that are marked incorrect by human verifiers are further refined by another set of annotators. Here we emulate the graph refinement phase with the help of a model. Specifically, our approach is a 2-stage pipeline -- first, an initial graph is generated by the baseline T5 model and second, an \textit{Explanation Graph Refinement} model conditions on the initial graph, along with the belief, argument and the stance to refine the graph. The refiner is also a T5 model fine-tuned with the prefix ``Refine the Explanation Graph for'' on all positive and negative graphs described in Sec.~\ref{sec:data_augmentation}. Note that our approach differs from the actual data collection process in two aspects. Unlike the human-annotated graphs, which are refined only for semantic correctness, the model-generated graphs can be both structurally and semantically incorrect. Second, our approach does not involve a graph verification stage and thus, the refiner model acts on all (correct and incorrect) graphs generated in stage 1 and is thus trained with both correct and incorrect graphs.

\begin{table*}[t]
\small
\centering
\begin{tabular}{lrrrrrr}
\toprule
& SA$\uparrow$ & StCA$\uparrow$ & SeCA$\uparrow$ & G-BS$\uparrow$ & GED$\downarrow$ & EA$\uparrow$ \\ \midrule
T5-Base \cite{saha2021explagraphs} & \textbf{87.2} & 38.7 & 19.0 & 33.6 & 0.71 & 20.8  \\
T5-Large & \textbf{87.2} & 51.0 & 34.7 & 43.9 & 0.61 & 29.5 \\ \midrule
Generate \& Refine & \textbf{87.2} & 52.5 & 37.7 & 45.3 & 0.60 & 30.0 \\
Pos Data Aug & \textbf{87.2} & 54.5 & 41.5 & 46.9 & 0.58 & 30.2 \\
Max-Margin & \textbf{87.2} & 56.7 & \textbf{43.5} & 48.6 & 0.57 & 30.5 \\
Contrastive & \textbf{87.2} & \textbf{60.5} & 42.5 & \textbf{52.1} & \textbf{0.52} & \textbf{33.1} \\
\midrule
Upper Bound & 91.0 & 91.0 & 83.5 & 71.1 & 0.38 & 46.8
\\\bottomrule                                             
\end{tabular}
\vspace{-5pt}
\caption{Comparison of all models across all metrics on the ExplaGraphs \cite{saha2021explagraphs} test set. \new{Improvement in SeCA is statistically significant (computed using Bootstrap test \cite{efron1994introduction}) with $p < 0.005$.}
}
\vspace{-12pt}
\label{tab:model_test}
\end{table*}

\subsection{Contrastive Graph Generation Model}
\label{sec:contrastive}
Our Contrastive Graph Generation Model (Fig.~\ref{fig:data_aug}) also leverages both positive and negative graphs but instead of doing so in a 2-stage \textit{Generate \& Refine} model, uses a contrastive learning framework \cite{khosla2020supervised, gunel2020supervised}. Given a ground-truth graph $\mathcal{G}^{(g)}$, a positive graph $\mathcal{G}^{(p)}$ and a set of negative graphs $\{\mathcal{G}^{(n)}_i\}_{i=1}^{M}$, contrastive learning aims to learn the graph representations such that the gold graph's representation is close to that of the synthetic positive graph while being distant from those of the negative graphs. Similar to \citet{cao2021cliff}, we use the last layer of the decoder in T5 as the representation of each token in the graph and obtain the graph representation by averaging over the constituent token representations. Let the graph representations be denoted by $h^{(g)}$, $h^{(p)}$ and $\{h^{(n)}_i\}_{i=1}^{M}$. Given $ \mathcal{H}^{(g)} = \{h^{(p)}\} \bigcup \{h^{(n)}_i\}_{i=1}^{M}$, our overall loss combines the cross-entropy loss $\mathcal{L}_{\mathit{CE}}$ and the InfoNCE contrastive loss \cite{Oord2018RepresentationLW} $\mathcal{L}_{\mathit{CL}}$ as shown below.
\begin{eqnarray}
    \mathcal{L}_{\mathit{CL}} = - \log \frac{\exp(\simi(h^{(g)},h^{(p)}) / \tau)}{\sum_{h_i \in \mathcal{H}^{(g)}} \exp(\simi(h^{(g)},h_i) / \tau)} \nonumber\\
    \mathcal{L} = \mathcal{L}_{\mathit{CE}} + \alpha \mathcal{L}_{\mathit{CL}} \nonumber
\end{eqnarray}
where $\alpha$ and the temperature $\tau$ are the hyperparameters and $\simi()$ denotes the cosine similarity function between the graph representations.

\section{Experiments}

\subsection{Impact of Different Models on Graph Structural and Semantic Accuracy}

In Table~\ref{tab:model_test}, we compare the various modeling techniques described in Sec.~\ref{sec:model} and their effect on the structural and semantic correctness of the generated graphs. While our primary metrics of interest are Graph Structural Accuracy (StCA) and Semantic Accuracy (SeCA), following prior work \cite{saha2021explagraphs}, we also report Stance Accuracy (SA), Graph-BertScore (G-BS), Graph Edit Distance (GED) and Edge Accuracy (EA). 

\paragraph{Effect of Model Size and Training Data. }The T5-Large model uses the same setup as the T5-Base model experimented with in \citet{saha2021explagraphs}. We observe that using a larger T5 model improves StCA by 12\% and SeCA by 16\%. This finding is in line with other commonsense reasoning tasks \cite{lourie2021unicorn, elazar2021back} which also show that fine-tuning a larger language model typically leads to better performance. Together with the results reported in Table \ref{tab:data_size}, we conclude that much of the improvement in explanation graph generation comes from increasing the training data and using a larger model. Given its superior performance, we build our proposed models on T5-large. 
\paragraph{Results with Generate \& Refine Model. }The \textit{Generate \& Refine} model (Sec.~\ref{sec:gen-refine})  improves all metrics; however the gains are small. Note that this model refines all graphs (correct or not) and can lead to already correct graphs becoming incorrect after refinement. In practice, we observe that most graphs do not change much after refinement which we believe stems from the model's inability to distinguish between correct and incorrect graphs.

\paragraph{Effect of Positive Graph Perturbations.} On retraining T5 augmented with the positively perturbed graphs (Sec.~\ref{sec:positive}), we observe that it obtains significant improvement over T5 and \textit{Generate \& Refine} both in structural and semantic accuracy. Note that, by construction, the positive graphs only differ in the commonsense concepts (not part of the belief or argument) while keeping the structure intact. Hence, the model has more supervision about the semantics of the graphs as opposed to the structural constraints. This is reflected in the larger improvement in SeCA. The positive graphs, being structurally correct, also reinforces the model's belief about structural correlation with correct graphs, thus leading to some improvement in StCA as well.

\paragraph{Effect of Negative Graph Perturbations. }The \textit{Max-Margin} model (Sec.~\ref{sec:max-margin}) leverages all structurally and semantically incorrect graphs and obtains up to 6\% and 9\% improvement in StCA and SeCA respectively over the baseline T5 model. The model implicitly learns the structural constraints through relevant supervision and the margin-based loss enables it to learn a better boundary between correct and incorrect graphs. Similarly, the semantically perturbed graphs improves the model's relation prediction capability between concepts. The \textit{Max-Margin} model outperforms the \textit{Pos Data Aug} model because of the former having access to both structural and semantic supervision while the latter is only augmented with structurally similar graphs.

\paragraph{Effect of Positive and Negative Graph Perturbations with Contrastive Learning.} The \textit{Contrastive Graph Generation} model (Sec.~\ref{sec:contrastive}) leverages both positive and negative graphs and improves StCA to 60\% with comparable SeCA to the \textit{Max-Margin} model. The overall improvements in StCA and SeCA are 9\% and 8\% respectively compared to T5. We hypothesize that the constrastive model does not lead to further improvement in SeCA because of the structurally similar positive graphs. This can potentially be improved by incorporating more structurally diverse graphs. Finally, our best SeCA is far from perfect and significant future work can be done in improving the graph semantics. Further ablations of negative graphs and human evaluation are done on the \textit{Max-Margin} model, due to its slightly higher SeCA.

\begin{table}[t]
\small
\centering
\resizebox{\columnwidth}{!}{
\begin{tabular}{lrrrrr}
\toprule
& StCA$\uparrow$ & SeCA$\uparrow$ & G-BS$\uparrow$ & GED$\downarrow$ & EA$\uparrow$ \\ \midrule
T5-Large & 46.5 & 31.6 & 36.8 & 0.66 & 26.7 \\
+ SySt & 50.2 & 34.1 & 40.7 & 0.64 & 27.4 \\
+ SySe & 50.7 & 35.1 & 40.8 & 0.63 & 27.3 \\
+ HuSe & 49.5 & 38.4 & 39.4 & 0.64 & 26.1 \\ 
\bottomrule                   
\vspace{-10pt}       
\end{tabular}
}
\caption{Ablation study showing the effect of different types of negative graphs on ExplaGraphs dev set.}
\vspace{-5pt}
\label{tab:effect_aug}
\end{table}

\begin{table}[t]
\small
    \centering
    \begin{tabular}{lrrr}
    \toprule
         & Valid$\uparrow$ & StCA$\uparrow$ & G-BS$\uparrow$ \\ \midrule
         T5-Base & 88.8 & 88.7 & 54.4  \\
         Max-Margin & 89.1 & 87.7 & 55.7 \\
         Contrastive & \textbf{97.5} & \textbf{96.9} & \textbf{57.2} \\ \bottomrule
    \end{tabular}
    \caption{Comparison of T5, Max-Margin and Contrastive models for temporal graph generation.}
    \label{tab:temporal}
    \vspace{-15pt}
\end{table}

\vspace{-2pt}
\subsection{Human Evaluation of Graph Semantics}
\vspace{-1pt}
Automatically evaluating graphs for semantic correctness is challenging. We conduct human evaluation to further validate our findings. We compare the graphs generated by T5 and our \textit{Max-Margin} model on Amazon Mechanical Turk where three annotators choose which graph is better or if they are mostly similar (instructions in Appendix~\ref{appendix:human_eval}). For fair comparison, we evaluate only those samples where both models predict the correct stance and the graphs are also structurally correct. In fact, this lets us evaluate the semantic aspect in isolation when both graphs are structurally correct. With majority voting on 150 samples, we observe that our \textit{Max-Margin} model's graphs are preferred 13\% more times compared to those of the T5 model (43\% vs 30\% and statistically significant with $p$ < 0.05) while in 22\% cases, the graphs are marked similar (remaining have no majority).

\vspace{-2pt}
\subsection{Ablation with Negative Graphs} 
\vspace{-1pt}
In Table~\ref{tab:effect_aug}, we show the effect of different types of negative graphs. We compare the results on the ExplaGraphs validation set by leveraging Synthetic Structural (SySt), Synthetic Semantic (SySe) and Human-created Semantic (HuSe) graphs with the \textit{Max-Margin} graph generation model. All types of negatives graphs lead to consistent increase in SeCA. Leveraging human-created negative graphs leads to a bigger gain in SeCA because of the hardness and diversity in these graphs \new{and hence are the best candidates for contrastive learning.}

\subsection{Generalization to Other Graph Generation Tasks}
\label{sec:temporal}
We test the generalizability of constructing structurally and semantically perturbed graphs for contrastive learning by also experimenting on a temporal graph generation task \cite{madaan2021neural} that requires constructing a temporal graph from a document. The nodes in the graph are events from the document and the edges are temporal relations between events (``before'', ``after'', etc). Following our overall goal of improving graph generation with limited data, we randomly sample 1.3\% of the overall corpus ($\sim$9.5k samples) as the training data such that all graphs are connected DAGs. Similar to ExplaGraphs, we create structurally negative graphs with disconnected and cyclic graphs and semantic negative graphs by perturbating the temporal relations. E.g., if an edge relation is ``before'', we replace it with ``after''. We construct positive graphs by replacing edges like ``A before B'' with ``B after A'' (more details in Appendix~\ref{appendix:temporal}). In Table \ref{tab:temporal}, we report structural correctness accuracy (StCA) (percentage of connected DAGs) and Graph-BertScore (G-BS) for measuring approximate semantic correctness wrt gold graphs. We observe that our contrastive model not only generates more valid graph encodings but also improves StCA by 8\% and G-BS by 3\%. 

\begin{table*}[t]
\small
\centering
\begin{tabular}{lrrrrr}
\toprule
& StCA$\uparrow$ & SeCA$\uparrow$ & G-BS$\uparrow$ & GED$\downarrow$ & EA$\uparrow$ \\ \midrule
SySt + SySe + HuSe & 49.5 & 38.4 & 39.4 & 0.64 & 26.1 \\ 
SySt + SySe + HuSe + \textit{HuSe-Gen (IP)} & 53.5 & 38.7 & 42.1 & 0.62 & 28.1 \\
SySt + SySe + HuSe + \textit{HuSe-Gen (AE)} & 52.0 & 40.2 & 41.3 & 0.62 & 28.2 \\
\bottomrule                   
\vspace{-15pt}       
\end{tabular}
\caption{Effect of training the Max-Margin model with additional Human-like Semantic Negative Graphs on ExplaGraphs dev set. IP and AE refer to the two thresholding techniques for filtering generated negatives.}
\vspace{-12pt}
\label{tab:effect_huse_gen}
\end{table*}

\begin{figure}[t]
    \centering
    \includegraphics[width=0.95\columnwidth]{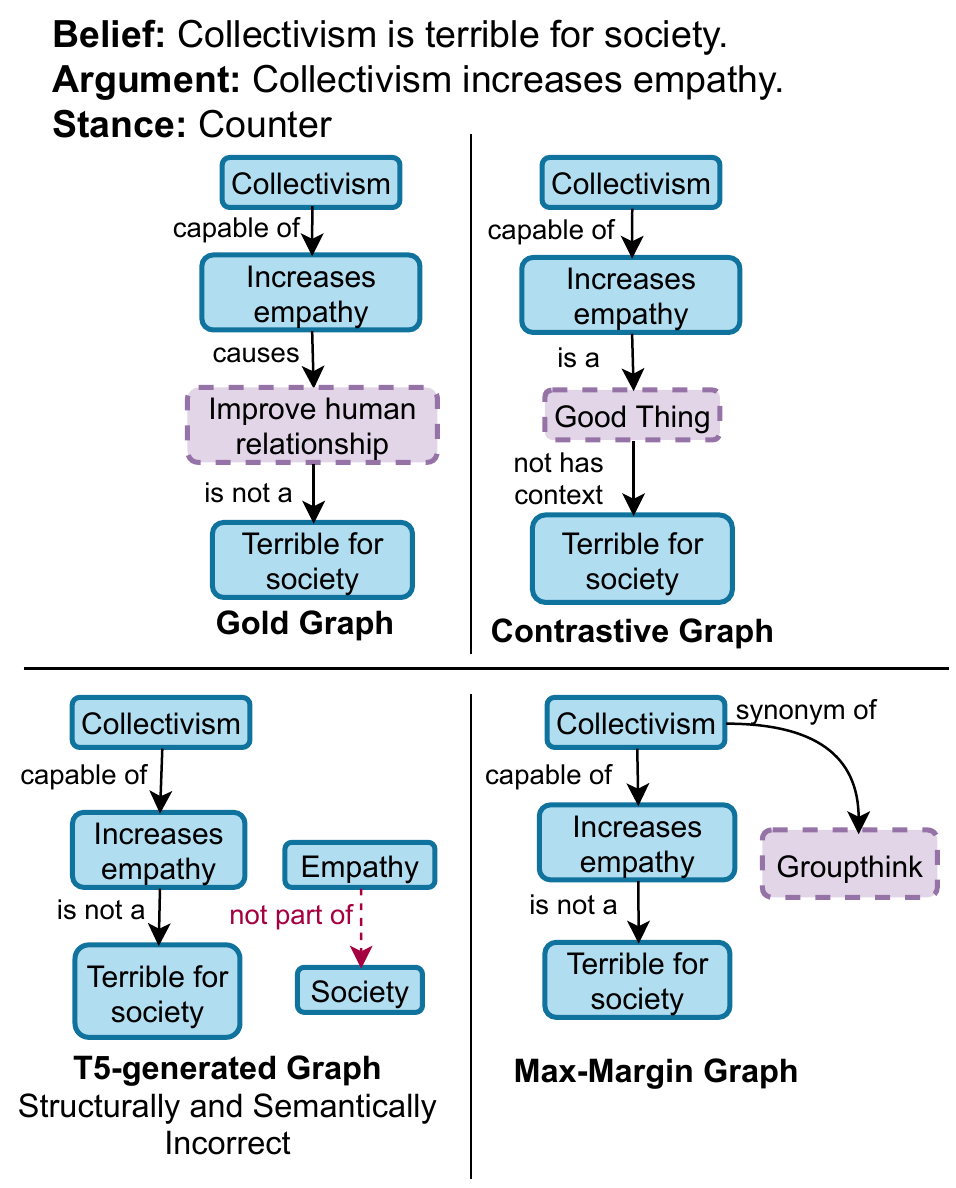}
    \vspace{-5pt}
    \caption{Qualitative analysis of explanation graphs.}
    \label{fig:qual1}
    \vspace{-15pt}
\end{figure}

\subsection{Analysis of Generated Graphs}
Fig.~\ref{fig:qual1} shows an example of the graphs generated by different models (more examples in Appendix~\ref{appendix:human_eval}). Unlike T5, our models' graphs are both structurally and semantically correct with diverse commonsense nodes (``Groupthink'', ``Good Thing''). While our models generate more correct graphs, they lack in structural diversity -- the Contrastive model generates 77\% of linear graphs (i.e., the nodes are in a linear chain) which is comparable to 75\% in the T5 model. This can be attributed to our structurally similar positive graphs as the model does not obtain enough supervision to generate diverse graphs. Structural diversity is not a measure of graph correctness; however, like diverse text generation \cite{vijayakumar2016diverse}, generating diverse graphs is an interesting direction for future work.

\subsection{Generating Human-like Semantic Negatives (HuSe-Gen)}

\new{In ExplaGraphs, human-created negatives account for 38\% of the samples for which the initially constructed graph was incorrect and was refined. Moreover, we see in the previous section that human-error graphs are the best negative candidates for contrastive learning (which is intuitive since tricky and subtle errors made by expert human annotators would make for some of the hardest negatives/distractors for a contrastive learning model to learn from). Hence, in this final section, we further explore whether it is also possible to automatically imitate and generate more of such \textit{harder human-like} incorrect graphs for the remaining samples as well. Our method consists of the following steps.}

\noindent \new{\textbf{Human-like Negative Edge Generation. }We first fine-tune a T5 model that conditions on the belief, argument and the stance to generate a set of incorrect edges (which is the set of edges that are present in the incorrect graph and not in the refined graph)}.

\noindent \new{\textbf{Human-like Negative Graph Construction. }This generated set of incorrect edges is then added to the correct graph to construct the incorrect graph, such that it is structurally correct and hence representative of human-like erroneous graphs.}

\noindent \new{\textbf{Filtering High-quality Negative Graphs.} Contrastive models will only benefit from these negatives if the negative edge generation model is accurate and generates edges that are actually incorrect. Hence, we control the quality of the generated incorrect graphs by the following two techniques -- (a) \textit{Thresholding via fraction of Acceptable Edges (AE):} We say that a generated incorrect edge is acceptable if it is not part of the correct graph and can be added to the correct graph without violating any structural constraints. We compute the fraction of acceptable edges for every generated negative graph and choose only those graphs with AE above a certain threshold $\delta$. Intuitively, this ensures that a high fraction of the generated edges are actually incorrect and hence when added to the correct graph, will lead to a sufficiently different (human-like) incorrect graph. (b) \textit{Thresholding via Incorrect Probability of a graph (IP):} We use our SeCA metric model (that classifies a graph into support, counter, or incorrect class) to compute the probability of the generated graph being incorrect and choose those graphs that are above a certain threshold $\gamma$ of incorrect probability.}

\new{We set $\delta=0.4$ and $\gamma=0.5$ (tuned on the dev set) and train the Max-margin model using these additionally generated human-like negative graphs. As shown in Table~\ref{tab:effect_huse_gen} both thresholding approaches lead to further improvements over using just the human-created negative graphs. These initial promising results for emulating hard/tricky human errors as strong negatives for contrastive learning will hopefully lead to further future work in this interesting direction.}

\vspace{-3pt}
\section{Conclusion}
\vspace{-2pt}

We presented an empirical study of graph structure and semantics for end-to-end explanation graph generation from pre-trained language models and showed that the generated graphs often violate structural constraints or are semantically incorrect. We significantly improve both the structural and semantic accuracy of graph generation by proposing contrastive learning models that leverage simple yet efficient methods of graph perturbations and also generalize to similar graph generation tasks.

\section*{Ethical Considerations}   

From an ethics standpoint, we provide a brief overview and show samples from the datasets that our models are trained on throughout the paper and also in the Appendix. Explanation graph generation improves the interpretability of neural commonsense reasoning systems and could prove to be effective in understanding and debugging such models. Hence we do not foresee any major risks or negative societal impact of our work. However, like any other ML model, the graphs generated by our models may not always be completely accurate and hence should be used with caution for real-world applications.

\section*{Acknowledgements}
\new{We thank the reviewers for their helpful feedback and the annotators for their time and effort. This work was supported by DARPA MCS
Grant N66001-19-2-4031, NSF-CAREER Award 1846185, DARPA YFA17-D17AP00022, ONR Grant N00014-18-1-2871, Microsoft Investigator Fellowship, and Munroe \& Rebecca Cobey Fellowship. The views in this article are those of the
authors and not the funding agency.}    

\bibliography{custom}
\bibliographystyle{acl_natbib}

\appendix
\section{Evaluation Metrics for ExplaGraphs}
\label{appendix:metrics}
Below we provide brief descriptions of the evaluation metrics used for the ExplaGraphs task. For further details, we refer readers to prior work \cite{saha2021explagraphs}.

\paragraph{Structural Correctness Accuracy of Graphs (StCA).} It computes the fraction of graphs where all the structural constraints are satisfied.

\paragraph{Semantic Correctness Accuracy of Graphs (SeCA).} SeCA is a model-based metric that computes the fraction of graphs that are both structurally and semantically correct. For computing SeCA, prior work trains a 3-way RoBERTa \cite{liu2019roberta} classifier that given a belief and a generated explanation graph, infers whether the graph supports the belief, counters the belief or is incorrect (because of incoherent edges). If it predicts support or counter and this stance matches the gold stance, then the graph is considered semantically correct. In essense, SeCA works on the principle that an explanation graph is semantically correct if a stance can be unambiguously inferred from it (by a model in this case or a human) and that stance is the same as the gold stance. Note that SeCA is a reference-free metric (does not use the ground-truth graph) and hence is invariant to structural variations in explanation graphs. 

\paragraph{Graph-BertScore (G-BS).} Graph-BertScore is an extension of BertScore \cite{bertscore} for computing the degree of match between the predicted graphs and the ground-truth graphs. It treats a graph as a set of edges and computes the best match between the gold edges and the predicted edges, where the matching score between a pair of edges is given by the BertScore F1.

\paragraph{Graph Edit Distance (GED).} GED is the standard Graph Edit Distance for graphs, measuring the number of edit operations (addition, deletion, and replacement of nodes and edges) to transform one graph to the other and further normalized by an appropriate normalizing constant.

\paragraph{Edge Accuracy (EA).} The final metric, Edge Accuracy (EA) measures the fraction of edges in the graph that are important. An edge is considered important if removing it from the graph leads to a drop in the gold stance prediction confidence.

\begin{table}[]
\centering
\small
\begin{tabular}{cccc}
\toprule
SySt & SySe & HuSe & Total \\ \midrule
7522 & 2368 & 1336 & 11226 \\ \bottomrule
\vspace{-15pt}
\end{tabular}
\caption{\label{tab:neg-data} Count of negative graphs in each category.}
\vspace{-10pt}
\end{table}

\section{Statistics of Graph Perturbations}

We create a total of 11k negative graphs. Table \ref{tab:neg-data} shows the respective counts of the negative graphs belonging to synthetic structural (SySt), synthetic semantic (SySe) and human-created semantic (HuSe) categories. 

\section{Temporal Graph Generation}
\label{appendix:temporal}
\begin{figure*}[t]
    \centering
    \includegraphics[width=\textwidth]{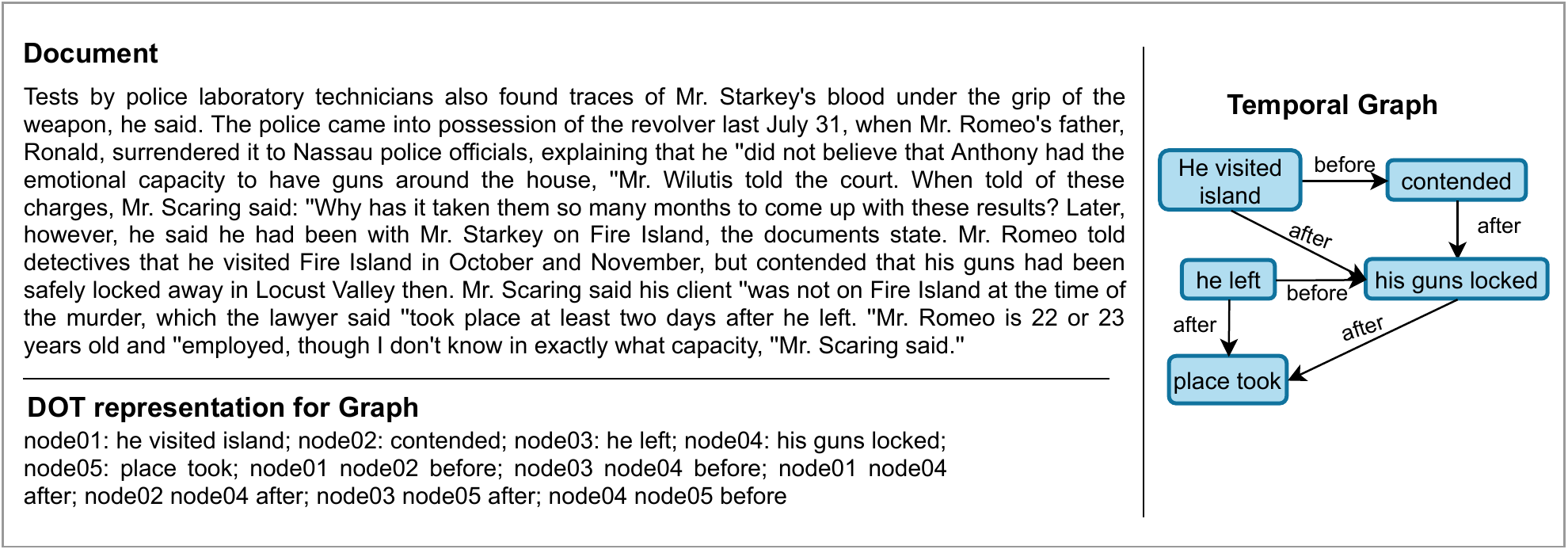}
    \caption{An example of the Temporal Graph Generation Task \cite{madaan2020eigen} showing the source document, the target temporal graph and the corresponding DOT representation.}
    \label{fig:temporal_example}
\end{figure*}

The task of temporal graph generation requires constructing a temporal graph from a document (see Fig. \ref{fig:temporal_example}). The nodes in the graph are events from the document (e.g., ``Markovic jailed'' or ``Covering up attempted murder'') and the edges are temporal relations between the events (e.g., ``Markovic jailed; before; Covering up attempted murder''). The authors consider five temporal relations (``before'', ``after'', ``simultaneous'', ``is included'' and ``includes'') and build an automatically constructed large-scale dataset for the task.  Following our overall goal of improving graph generation in limited data settings, we randomly sample 1.3\% of the overall corpus ($\sim$ 9.5k samples) as the training corpus such that all graphs are connected DAGs.\footnote{Since the dataset was constructed automatically, we found about 10\% of the graphs to be disconnected or cyclic.} Following~\citet{madaan2021neural}, we represent graphs in DOT format~\cite{koutsofios1996drawing} as shown in Fig.~\ref{fig:temporal_example}. We find that the specifics of the graph representations do not matter much, as long as all the edges are concatenated in one particular ordering (either DFS, BFS or Topological order).

We construct semantic negative graphs by randomly sampling a fraction of the edges and performing the following operations. If an edge relation is one of ``before'', ``after'' or ``simulatenous'', we replace it with any other relation from this set and if the relation is one of ``is included'' or ``includes'' we replace it with the other relation. Note that these perturbations will always lead to incorrect graphs because ``A before B'' implies that ``A after B'' or ``A simultaneous B'' do not hold. Finally, we construct positive graphs by randomly sampling a fraction of edges and replacing them using the following rules: (1) ``A before B'' with ``B after A'' and viseversa, (2) ``A simultaneous B'' with ``B simultaneous A'', (3) ``A includes B'' with ``B is included A''. Note that all these operations preserve the temporal meaning of the graph and are done in a way such that the perturbed graph continues to be a connected DAG.

\begin{table}[]
    \centering
    \begin{tabular}{llll}
    \toprule
        Dataset & Train & Dev & Test \\ \midrule
        ExplaGraphs & 2368 & 398 & 400 \\
        Temporal (Sampled) & 9531 & 953 & 949 \\\bottomrule
    \end{tabular}
    \caption{Train, validation and test split sizes of the two datasets. For Temporal Graph Generation, we randomly sample 1.3\% of the overall corpus \cite{madaan2021neural}.}
    \label{tab:dataset_stats}
\end{table}

\begin{table*}[t]
\small
\centering
\begin{tabular}{lrrrrrr}
\toprule
& SA$\uparrow$ & StCA$\uparrow$ & SeCA$\uparrow$ & G-BS$\uparrow$ & GED$\downarrow$ & EA$\uparrow$ \\ \midrule
T5-Base & \textbf{86.2} & 35.4 & 15.5 & 27.7 & 0.75 & 19.8 \\
T5-Large & \textbf{86.2} & 46.5 & 31.6 & 36.8 & 0.66 & 26.8 \\ \midrule
Generate \& Refine & \textbf{86.2} & 46.8 & 34.4 & 37.2 & 0.66 & 27.2 \\
Pos Data Aug & \textbf{86.2} & 50.0 & 37.6 & 39.6 & 0.64 & 28.4 \\
Max-margin & \textbf{86.2} & 49.5 & \textbf{38.4} & 39.4 & 0.64 & 26.1  \\
Contrastive & \textbf{86.2} & \textbf{52.7} & 37.9 & \textbf{41.7} & \textbf{0.62} & \textbf{29.8} \\
\bottomrule
\end{tabular}
\caption{Comparison of our models with baseline T5 models across all metrics on ExplaGraphs dev set.
}
\label{tab:model_dev}

\end{table*}

\begin{table*}[t]
\small
\centering
\begin{tabular}{lrrrrrr}
\toprule
& SA$\uparrow$ & StCA$\uparrow$ & SeCA$\uparrow$ & G-BS$\uparrow$ & GED$\downarrow$ & EA$\uparrow$ \\ \midrule
BART-Base & \textbf{87.2} & 25.7 & 13.0 & 22.0 & 0.81 & 12.8 \\
BART-Large & \textbf{87.2} & 34.2 & 22.2 & 28.9 & 0.75 & 20.0 \\
Contrastive & \textbf{87.2} & \textbf{40.7} & \textbf{26.3} & \textbf{31.3} & \textbf{0.71} & \textbf{22.3} \\
\bottomrule
\end{tabular}
\caption{Effect of Contrastive Learning with BART on ExplaGraphs test set.
}
\label{tab:bart}
\end{table*}

\section{Experimental Setup}
\label{appendix:setup}

Table \ref{tab:dataset_stats} shows the number of train, validation and test samples of the two datasets we experiment with. We build our models on top of the Hugging Face transformers library \cite{wolf2019huggingface}.\footnote{\url{https://github.com/huggingface/transformers}} All models for the ExplaGraphs dataset\footnote{\url{https://github.com/swarnaHub/ExplaGraphs}} \cite{saha2021explagraphs} are trained with a batch size of $8$ and an initial learning rate of $3*10^{-5}$ for a maximum of $15$ epochs. The maximum input and output sequence lengths are both set to $150$. For the max-margin graph generation model, we set both the hyperparameters $\alpha$ (mixing ratio) and $\beta$ (margin) to $1.0$ while for the contrastive graph generation model, we set $\alpha$ to $0.1$. For the temporal graph generation task\footnote{\url{https://github.com/madaan/temporal-graph-gen}} \cite{madaan2021neural}, we train all models with a batch size of $4$ and an initial learning rate of $3*10^{-5}$ for a maximum of $10$ epochs. The maximum input and output sequence lengths are set to $512$ and $256$ respectively. On this task, the hyperparameters $\alpha$ and $\beta$ for the max-margin model are again set to $1.0$ while for the contrastive graph generation model, we set $\alpha$ to $0.2$. 

Across all models and tasks, graphs are generated using beam search decoding with a beam size of $4$. The batch size and learning rate are manually tuned in the range $\{4, 8, 16\}$ and \{$10^{-5}$, $2*10^{-5}$, $3*10^{-5}$\} respectively and the best models are chosen based on the respective validation set performance. Similarly, the mixing ratio hyperparameter $\alpha$ is manually tuned in the range $\{0.1, 0.2, 0.5, 1.0\}$. The random seed is set to $42$ in all our experiments. The total number of parameters in our models is similar to T5-Base (220M) or T5-Large (770M) depending on the base architecture. All our experiments are executed on a single A100 Nvidia GPU. Each epoch of the contrastive model has an average runtime of 30 mins for ExplaGraphs and 2.5 hours for Temporal Graph Generation.

\begin{figure*}[t]
    \centering
    \includegraphics[width=\textwidth]{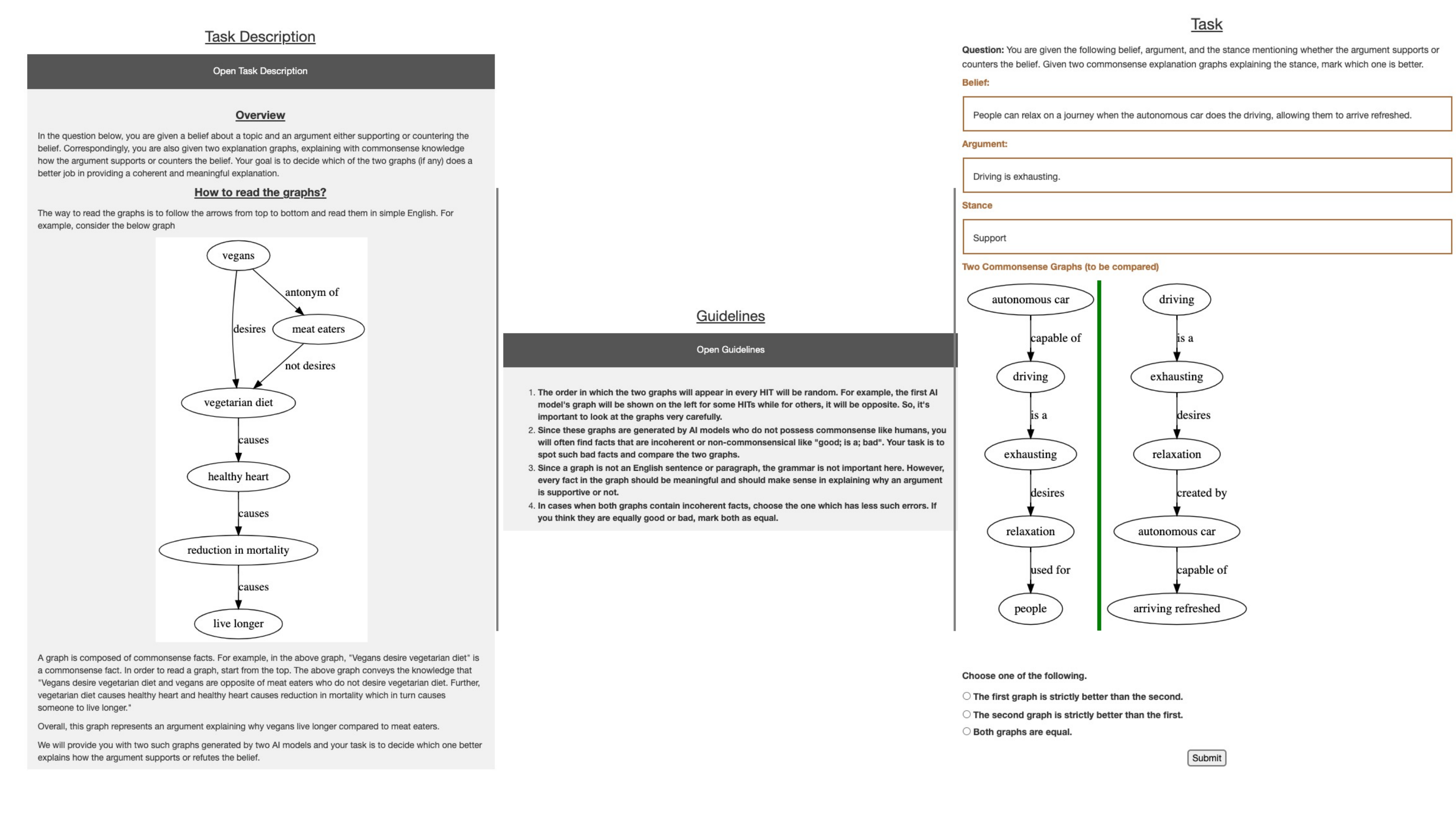}
    \caption{Interface for human evaluation of commonsense explanation graphs.}
    \label{fig:template}
\end{figure*}

\begin{figure*}[t]
    \centering
    \includegraphics[width=\textwidth]{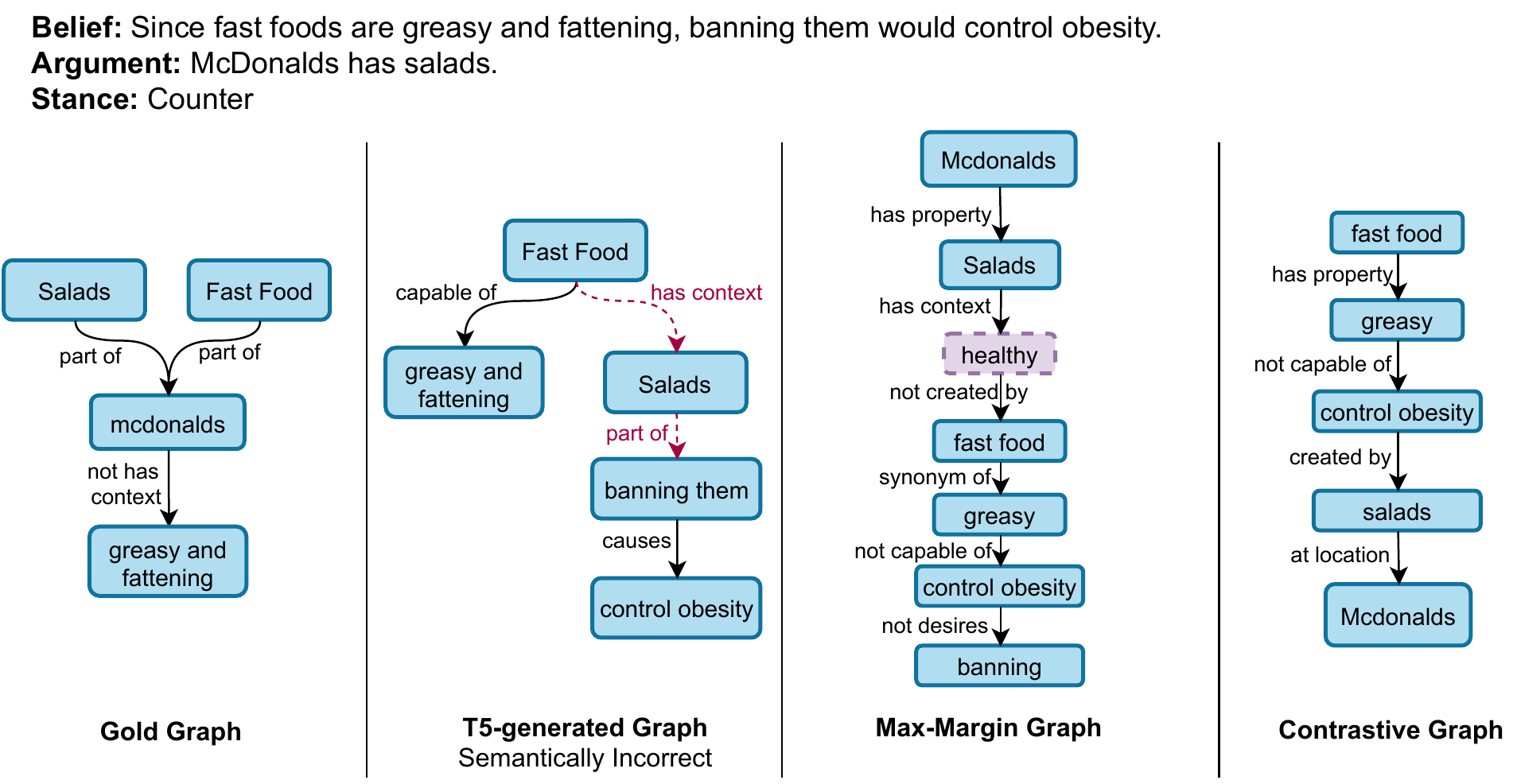}
    \caption{Example of explanation graphs generated by different models. The baseline T5-generated graph is semantically incorrect (incoherent relations marked in dashed red) while our proposed models generate both structurally and semantically correct graphs.}
    \label{fig:qual2}
\end{figure*}

\section{Results}
\label{appendix:results}
Table \ref{tab:model_dev} shows the results of all models on the ExplaGraphs \cite{saha2021explagraphs} validation set.

\new{\paragraph{Experiments with BART.} In Table~\ref{tab:bart}, we show the performance of BART~\cite{lewis2020bart} on ExplaGraphs~\cite{saha2021explagraphs} test set. Unsurprisingly, a larger BART model obtains a much higher StCA and SeCA compared to BART-Base. However, we find T5 to perform much better on this task. Applying contrastive learning on top of BART leads to improvements across all metrics, thereby showing our method's generalizability across different pre-trained language models.}

\begin{table}[t]
\small
\centering
\resizebox{\columnwidth}{!}{
\begin{tabular}{lrrrrr}
\toprule
& StCA$\uparrow$ & SeCA$\uparrow$ & G-BS$\uparrow$ & GED$\downarrow$ & EA$\uparrow$ \\ \midrule
Max-Margin & 56.7 & 43.5 & 48.6 & 0.57 & 30.5 \\
+ Atomic & \textbf{58.2} & \textbf{45.0} & \textbf{49.9} & \textbf{0.56} & \textbf{30.9} \\\bottomrule
\vspace{-10pt}  
\end{tabular}}
\caption{Effect of fine-tuning with additional commonsense knowledge from Atomic.}
\vspace{-5pt}  
\label{tab:effect_atomic}
\end{table}

\paragraph{Effect of Additional Commonsense Knowledge.} In Table~\ref{tab:effect_atomic}, we explore the impact of integrating additional commonsense knowledge to our \textit{Max-Margin} model. Specifically, we first fine-tune a T5 model on the facts based on ConceptNet relations from ATOMIC-2020~\cite{hwang2021comet}, a large-scale commonsense knowledge base. The fine-tuning objective is to predict the target concept given the source concept and the relation. Next, we fine-tune this model further on the end-task of graph generation which leads to small improvements in both StCA and SeCA. This suggests that better methods of inducing commonsense knowledge in these models can potentially lead to bigger gains with more semantically coherent graphs.

\section{Human Evaluation} 
\label{appendix:human_eval}
In Fig.~\ref{fig:template}, we show the interface for human verification of commonsense explanation graphs on Amazon Mechanical Turk. We select crowdworkers who are located in the US with a HIT approval rate higher than 96\%
and at least 1000 HITs approved. Since graph evaluation is a challenging task, we first explain how to read the graphs and also provide clear guidelines for comparing the quality of the two graphs.\footnote{The payment for each HIT is 0.25\$ at the rate of 12-15\$ per hour.}

\section{Examples of Generated Explanation Graphs}
\begin{figure*}[t]
    \centering
    \includegraphics[width=\textwidth]{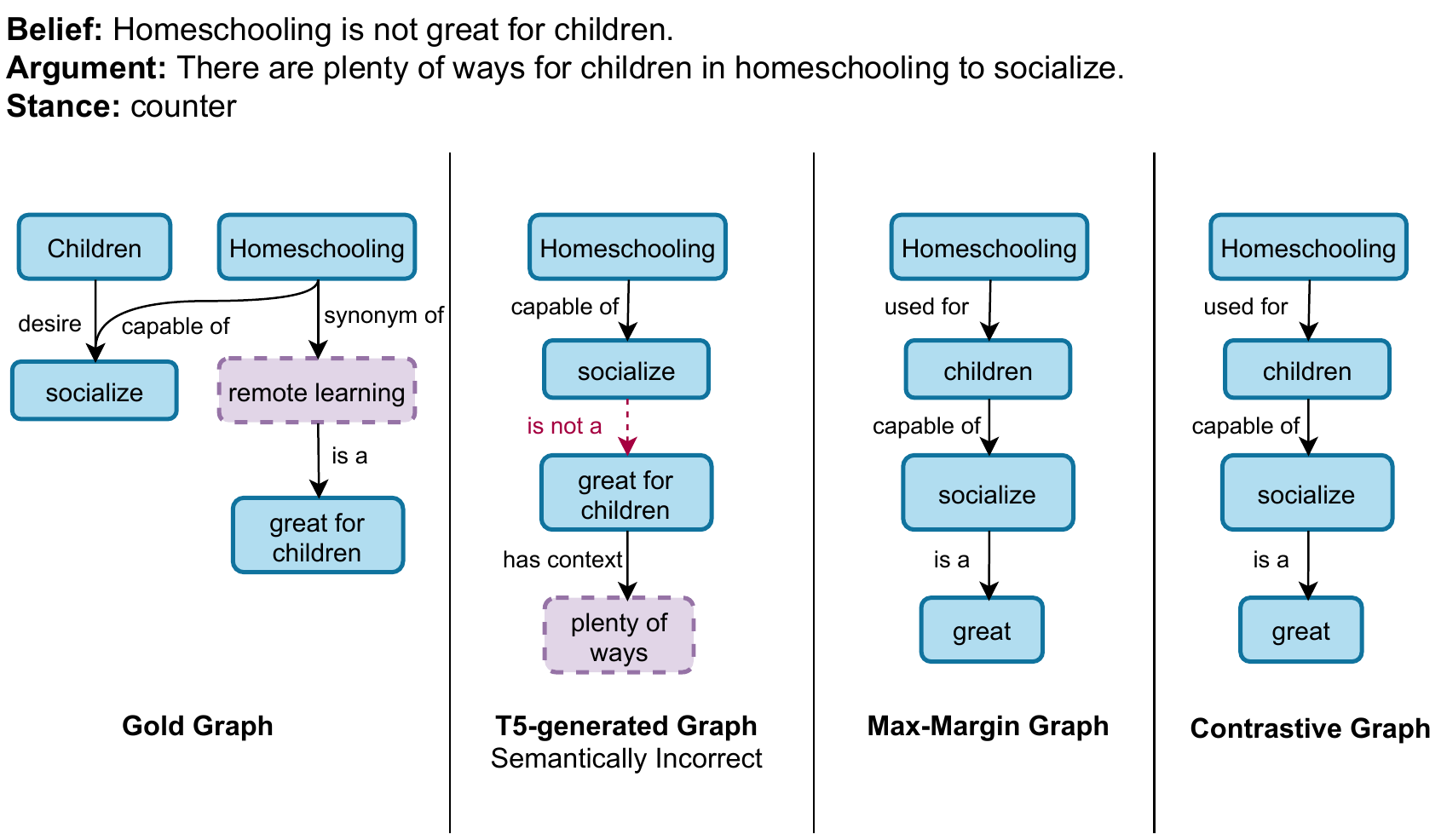}
    \caption{Example of explanation graphs generated by different models. The baseline T5-generated graph is semantically incorrect (incoherent relations marked in dashed red) while our proposed models generate both structurally and semantically correct graphs.}
    \label{fig:qual3}
\end{figure*}

\begin{figure*}[t]
    \centering
    \includegraphics[width=\textwidth]{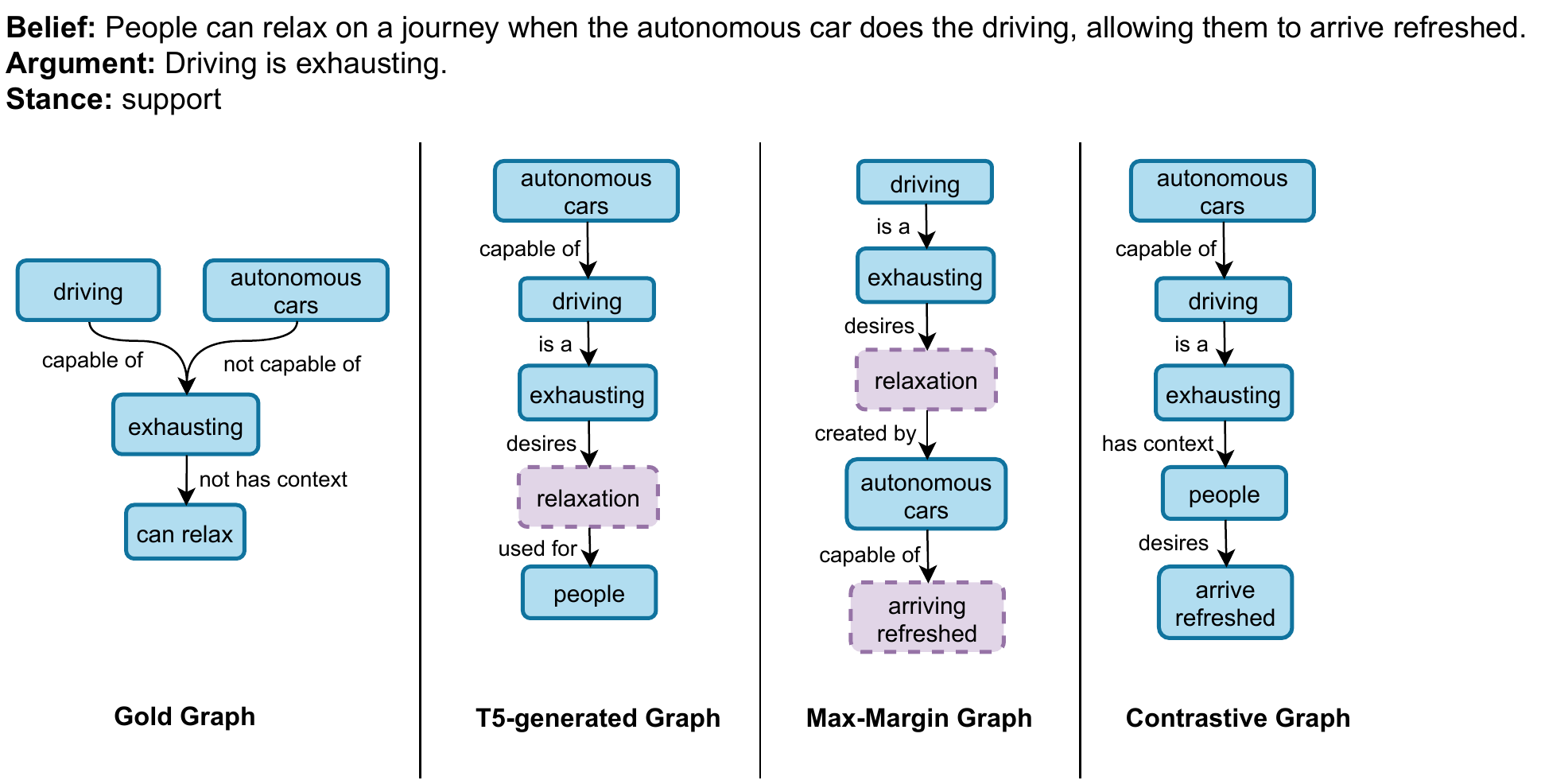}
    \caption{Example of explanation graphs generated by different models. All models generate structurally and semantically correct graphs while the individual nodes and edges differ.}
    \label{fig:qual4}
\end{figure*}

\begin{figure*}[t]
    \centering
    \includegraphics[width=\textwidth]{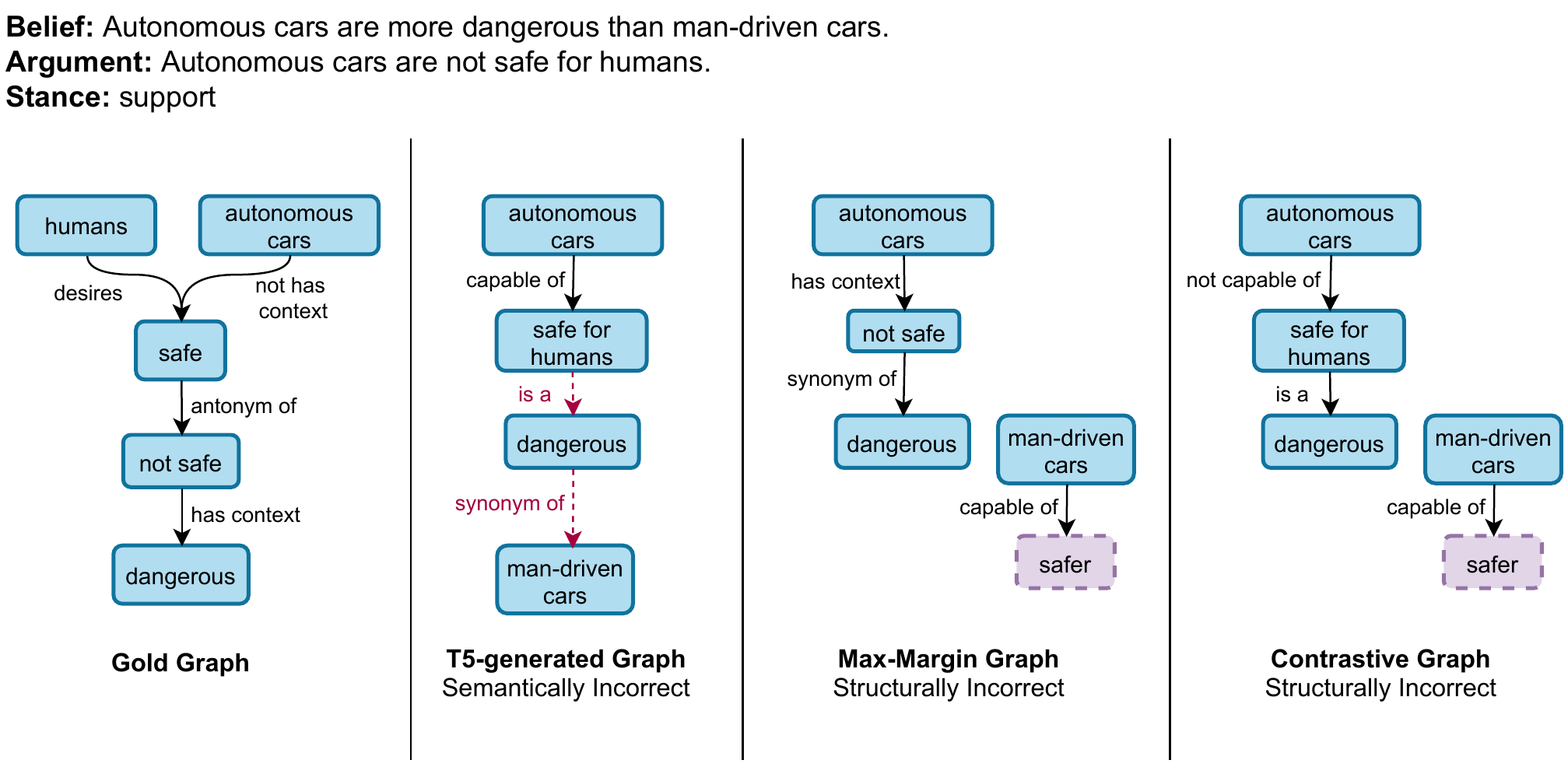}
    \caption{Example of explanation graphs generated by different models. T5 generates a semantically incorrect graph. Our models generate graphs, which while contain meaningful edges, are disconnected and hence are structurally incorrect.}
    \label{fig:qual5}
\end{figure*}

In Fig.~\ref{fig:qual2},~\ref{fig:qual3},~\ref{fig:qual4} and~\ref{fig:qual5}, we show various examples of explanation graphs generated by our models. In Fig. \ref{fig:qual2} and \ref{fig:qual3}, our proposed models improve upon the incorrect semantic relations from the T5 baseline graphs. Fig. \ref{fig:qual4} shows an example where all generated graphs, while different, are correct. Finally, Fig~\ref{fig:qual5} shows an example where although our proposed models improve the semantic aspect compared to the baseline graph, the generated graphs are disconnected and hence structurally incorrect. Overall, our quantitative results and human evaluation suggest that there is significant room for improvement on the task of commonsense explanation graph generation.

\end{document}